\documentclass[11pt]{article}

\usepackage[]{acl}

\usepackage{times}
\usepackage{latexsym}

\usepackage[T1]{fontenc}

\usepackage[utf8]{inputenc}

\usepackage{microtype}

\usepackage{url}
\usepackage[utf8]{inputenc}
\usepackage{color,soul}
\setul{0.45ex}{0.25ex}
\usepackage{graphicx}
\usepackage{subcaption}
\usepackage{amsmath}
\usepackage{booktabs}
\usepackage{multirow, makecell}
\usepackage{tikz}
\usepackage{graphicx}
\usepackage{array}
\usepackage{pgfplots}
\usepackage{algorithm}
\usepackage{textcomp}
\usepackage{xcolor}
\usepackage{times}
\usepackage{latexsym}
\usepackage{amsmath}
\usepackage{amsfonts}
\usepackage[noend]{algpseudocode}
\usepackage{lipsum}
\usepackage{multirow}
\usepackage{framed}
\usepackage{graphicx}
\usepackage{pifont}
\usepackage{longtable}
\usepackage{tikz}
\usepackage{booktabs}
\usepackage[all]{nowidow}
\usepackage{amssymb}
\usepackage{enumitem}
\usepackage{color,soul}
\usepackage{bm}
\usepackage{arydshln}
\usepackage{xspace}
\usepackage{siunitx}
\usepackage{outlines}
\usepackage{rotating}
\usepackage{subcaption}
\usepackage{ragged2e}

\pgfplotsset{compat=1.16}

\newif\ifcomments
\ifcomments
    \providecommand{\sameer}[2][]{{\protect\color{red}{[Sameer:\textbf{#1} #2]}}}
    \providecommand{\rob}[2][]{{\protect\color{violet}{[Rob:\textbf{#1} #2]}}}
    \providecommand{\alex}[2][]{{\protect\color{teal}{[Alex:\textbf{#1} #2]}}}
    \providecommand{\mingwei}[2][]{{\protect\color{blue}{[Ming-Wei:\textbf{#1} #2]}}}
\else
    \providecommand{\sameer}[2][]{}
    \providecommand{\rob}[2][]{}
    \providecommand{\alex}[2][]{}
    \providecommand{\mingwei}[2][]{}
\fi

\newcommand{\eat}[1]{}
\newcommand{\nlp}[1]{\texttt{\small #1}}
\newcommand{\taskacronym}{FRUIT\xspace}
\newcommand{\dataset}{\textsc{FRUIT-Wiki}\xspace}
\newcommand{\model}{\textsc{EdiT5}\xspace}
\newcommand{\src}{A^t}
\newcommand{\tgt}{A^{t'}}
\newcommand{\evidence}{\mathcal{E}^{t \rightarrow t'}}
\newcommand{\deltarouge}{UpdateROUGE\xspace}

\newcommand{\appendixprefix}{A}

\title{
\taskacronym \includegraphics[height=1em]{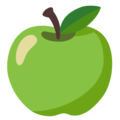}: Faithfully Reflecting Updated Information in Text
}

\author{
    \bf Robert L. Logan IV%
        \thanks{~~Work done during an internship at Google Research.}
        \thanks{~~Corresponding Author.}
        $^{1}$
        \hspace{0.3cm}
    \bf Alexandre Passos$^{2}$ \\
    \bf Sameer Singh$^{1}$ \hspace{0.3cm}
    \bf Ming-Wei Chang$^{\dagger 2}$ \\
    $^1$University of California, Irvine \hspace{0.3cm}
    $^2$Google Research \\
    \{\href{mailto:rlogan@uci.edu}{\tt rlogan},\href{mailto:sameer@uci.edu}{\tt sameer}\}\href{mailto:rlogan@uci.edu}{\tt @uci.edu} \\
    \{\href{mailto:apassos@google.com}{\tt apassos},\href{mailto:mingweichang@google.com}{\tt mingweichang}\}\href{mailto:mingweichang@google.com}{\tt @google.com}
}

\begin{document}
\maketitle
\begin{abstract}
    Textual knowledge bases such as Wikipedia require considerable 
    effort to keep up to date and consistent. 
    While automated writing assistants could potentially ease this burden,
    the problem of suggesting edits grounded in external knowledge has been under-explored.
    In this paper, we introduce the novel generation task of \emph{faithfully reflecting updated information in text} (\taskacronym) where the goal is to update an existing article given new evidence.
    We release the \dataset dataset, a collection of over 170K distantly supervised data produced from pairs of Wikipedia snapshots, along with our data generation pipeline and
    a gold evaluation set of 914 instances whose edits are guaranteed to be supported by the evidence.  
    We provide benchmark results for popular generation systems as well as \model---a T5-based approach tailored to editing we introduce that establishes the state of the art.
    Our analysis
    shows that developing models that can update articles faithfully requires new capabilities for neural generation models, and opens doors to many new applications.\footnote{Our data and code are available at: \\
{\small \url{https://github.com/google-research/language/tree/master/language/fruit}}.}

\end{abstract}

\section{Introduction}

\begin{figure*}[!t]
    \centering
    \includegraphics[width=\textwidth]{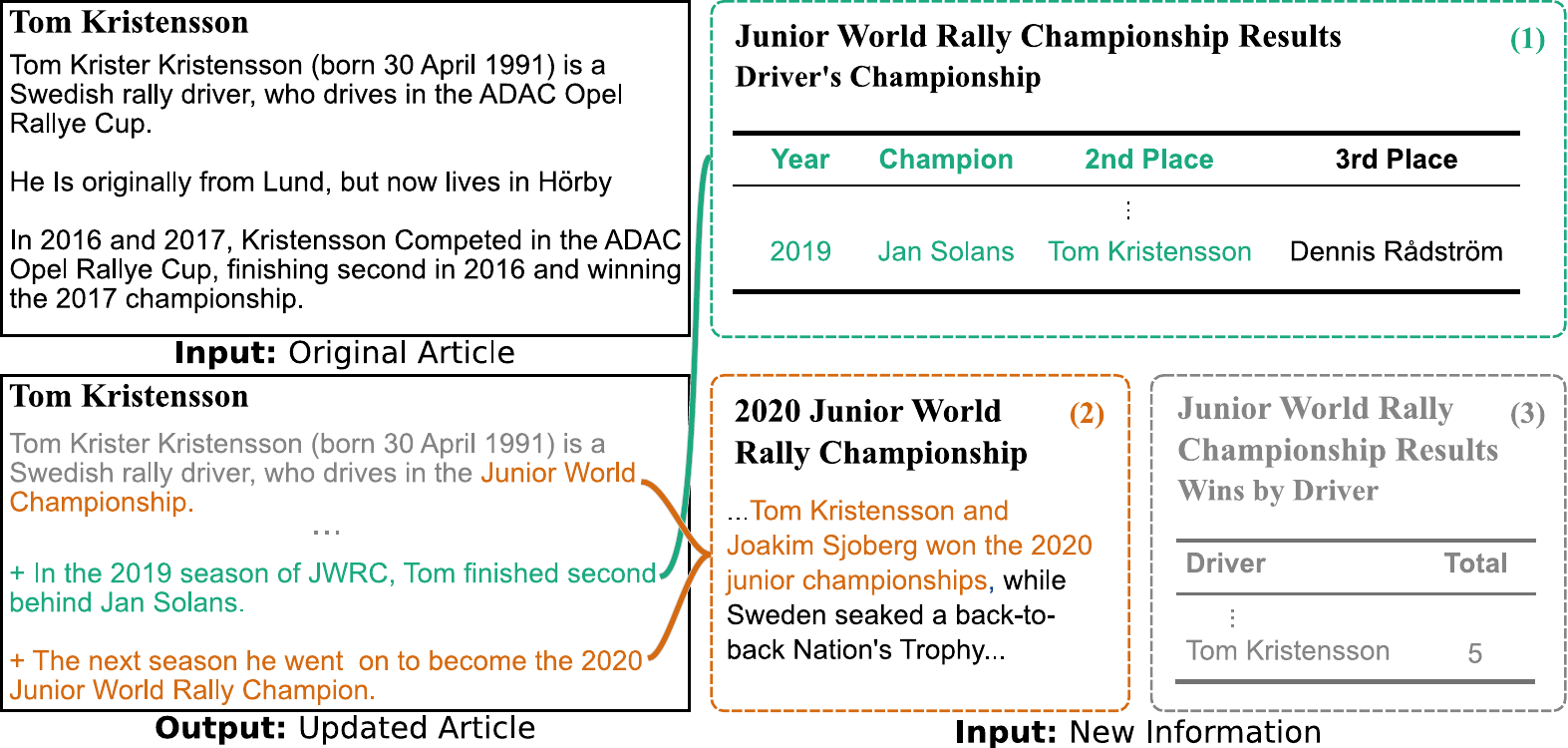}
    \caption{
    {\bf Illustration of the FRUIT task}.
    An outdated \emph{original article} and relevant \emph{new information} are provided as inputs, and the goal is to generate the {\em updated article}.
    In this example, the original article about Tom Kristensson was written in 2020, and 
    the new information is comprised of updated information about Tom Kristensson that has been added to other Wikipedia articles between 2020 and 2021.
    Given these inputs, the goal is to produce the updated 2021 version of article.
    Models need to identify the relevant supporting facts ({\color[HTML]{d56a11} orange} and {\color[HTML]{1ba97f} teal}) to generate faithful updates while ignoring superfluous information ({\color[HTML]{848484} grey}).
        }
    \label{fig:task-illustration}
\end{figure*}


Information changes on a constant basis.
Every day, athletes are traded to new teams, 
and musicians and actors produce new albums and TV shows.
Maintaining textual knowledge bases to keep track of these changes requires considerable community effort.
For instance, a team of 120K volunteer editors make 120 edits to English Wikipedia every minute, and write 600 new articles a day.\footnote{\url{https://en.wikipedia.org/wiki/Wikipedia:Statistics}}
As the knowledge base grows, the amount of maintenance effort is compounded by the need to keep the knowledge base consistent; e.g., each edit may render information in one of the existing 6.3M+ articles obsolete.

Assistive writing technologies have the potential to substantially reduce 
the burden of keeping text corpora up to date and consistent.
However, existing work has mainly focused on correcting grammar~\cite{Wang2020ACS}, reducing repetitive typing~\cite{chen2019gmail}, and following rhetorical directives~\cite{sun-etal-2021-iga}, whereas the problem of producing edits grounded in external knowledge has received little attention~\cite{kang-etal-2019-pomo}.
In contrast, numerous works have developed systems for distilling external knowledge into text (e.g., Wikipedia article generation) by treating the problem as multi-document summarization~\cite{Liu2018GeneratingWB,shi-etal-2021-descgen} or data-to-text generation~\cite{Bao2018TabletoTextDT,parikh-etal-2020-totto}.
However, these systems are not useful for updating existing texts as they can only generate text from scratch.
To help endow writing assistants with grounded editing capabilities,
we introduce the novel generation task of \emph{faithfully reflecting updated information in text} (\taskacronym), where the goal is to incorporate new information into an existing piece of text.
An illustration is provided in Figure~\ref{fig:task-illustration}.
Given an outdated Wikipedia article and collection of new information about the article's subject, \taskacronym requires updating the existing text so that it is consistent with the new information, as well as adding text to reflect new salient facts, e.g., in Figure~\ref{fig:task-illustration}, the first sentence is updated to reflect that Tom Kristensson now drives in the Junior World Championship, and new sentences are added to reflect his achievements in 2019 and 2020. 

FRUIT presents several unique challenges.
First, unlike many generation tasks, models cannot obtain good performance by solely relying on their parametric world knowledge.
Whenever the provided evidence contradicts parametric knowledge, the model must prefer the evidence, which recent work has shown is difficult for pretrained language models~\cite{krishna-etal-2021-hurdles,longpre-etal-2021-entity}.
Second, the generated text needs to be faithful to {\em both} the original article and the new evidence, 
\emph{except} when
evidence invalidates information in the existing article.
Finally, this task requires models to jointly read and analyze evidence from both textual and tabular sources and determine which is relevant and which can be ignored, thus combining challenging aspects of both multi-document summarization and data-to-text generation.

To facilitate research on this task, we release the \dataset dataset, a collection of over 170K distantly supervised (``\emph{silver}'') update-evidence pairs.
This dataset is produced by comparing pairs of English Wikipedia snapshots to identify
updates to an article between
two snapshots,
and associating information from the other articles that supports these updates under a distant supervision assumption.
As there is no guarantee that updates in the later Wikipedia snapshots can be supported by the collected evidence, we also collect a ``\emph{gold}'' evaluation set of 914 human annotated update-evidence pairs where unsupported claims have been removed without disturbing fluency.
We train and validate our models using silver data and then evaluate the final performance using gold data.

We establish initial benchmark results for a number of trivial and neural sequence-to-sequence baselines.
We also introduce \model, a T5-based model specially adapted for grounded editing, which establishes state-of-the-art performance on \dataset.
Through an extensive set of analyses, we identify a number of failure modes needed to be improved upon in order to obtain better performance on \dataset, as well as other interesting topics for future work on this task.
We additionally release our data collection pipeline to allow researchers to produce data from future Wikipedia snapshots and other languages, which we show to produce high-quality silver data.

\section{The \taskacronym Task}

\subsection{Task Definition}
\label{section:task-definition}

In this section we introduce the task of \emph{faithfully reflecting updated information in text} (\taskacronym).
Given an input piece of text focused on a topic or event, along with a collection of potentially new information about the subject of the text, the goal is to update the input text to reflect the new information.
A concrete illustration of the task is provided in Figure~\ref{fig:task-illustration}.
The original piece of text along with its updates are shown on the left, while the new information is shown on the right.

Formally, we assume access to pair of texts, $\src$ and $\tgt$, pertaining to a given subject, written at times $t$ and $t'$ (respectively).
In addition, we assume access to a set of new information, a.k.a., evidence, $\evidence = \left\{E_1, \ldots, E_{|\mathcal{E}|} \right\}$, mentioning the subject written between times $t$ and $t'$.
As is shown in Figure~\ref{fig:task-illustration}, the evidence can contain structured objects (e.g., excerpts from tables) as well as unstructured text.
Given $\src$ and $\evidence$ the goal is produce the updated text $\tgt$.

Successful completion of this task requires a number of complex and inter-related reasoning capabilities.
For one, models must be able to identify which evidence contradicts existing portions of the source article, and which evidence introduces new salient information about the subject in order to correctly choose whether to alter the existing text vs. add new text.
For example, in Figure~\ref{fig:task-illustration} the first sentence is updated to reflect that Tom Kristensson now races in a different competition, whereas new sentences are added describing his achievements in the years 2019 and 2020.
Models must also be able to determine whether a given piece of evidence should be used at all, i.e., perform content selection.
For example, in Figure~\ref{fig:task-illustration}, the number of rounds won by Kristennsen appears in the evidence but does not correspond to any piece of updated text.
Although some evidence may not appear in the updated article, the converse is not true, the system should aim to generate an updated article where all the updates are faithful to the evidence.

\subsection{Evaluation}
\label{section:evaluation}

In this section we introduce important considerations for evaluating \taskacronym systems.

\paragraph{Evaluate on Updated Text}
\mingwei{This subtitle is not very clear. Maybe "Evaluate on Updated Text"? Also it does not align very well wit the next item (Faithfulness). To align well, we should say Fluency or something. I do not have a solution yet}
\rob{What about "Don't overemphasize copied text"? I think clarity is more important than stylistic alignment (this subsection has 3 paragraphs and the other two headings aren't similar anyways).}
There is often considerable overlap between the original and updated text.
As we will see in Section~\ref{section:results-analysis} this poses a challenge for standard evaluation metrics like ROUGE~\cite{Lin2004ROUGEAP} as systems can achieve high scores without making any updates.
In this work, we propose to evaluate \taskacronym systems using an alternative metric, \deltarouge, that only considers updated sentences instead full texts.
For example, in Figure~\ref{fig:task-illustration}, the reference for \deltarouge only consists of the first and last two sentences.

\paragraph{Evaluate Faithfulness}

Ensuring that generations faithfully reflect information in the evidence and updated article is crucial.
However measuring faithfulness of generations is an active area of research~\cite{elikyilmaz2020EvaluationOT} and adapting existing metrics to the \taskacronym task is non-trivial.

As a simple proxy for faithfulness, we choose to measure the token overlap between named entities appearing in the generation and the target article/evidence, where entities are identified using the named entity recognizer used by \citet{guu2020realm} to perform salient span masking.
We specifically introduce the following measurements:
\begin{enumerate}[nosep,leftmargin=*]
    \item {\bf Unsupported Entity Tokens.} This metric shows the average number of entity tokens appearing in generated updates that do not appear in the \emph{source article} or \emph{evidence}. This is intended to capture the overall amount of unfaithful text, focusing on entities, where higher numbers indicate less faithfulness.
    \item {\bf Entity Precision and Recall.} These metrics capture the overlap of mentions between the generated updates and the \emph{target}. Entity precision measures the fraction of entity tokens appearing in the generated updates that appear in target entities, whereas
    entity recall measures the fraction of entity tokens in the target that appear in the entities in generated updates.
    The latter is similar to \deltarouge but only evaluated on entities, and thus, potentially less sensitive to paraphrasing.
\end{enumerate}

\paragraph{Parametric Knowledge Consideration}
\taskacronym systems should incorporate information from the provided evidence into the update, and not information that happened to be present during training or pretraining.
In this work we attempt to address this by evaluating models only on updates that were made to the text after the data used to pretrain and finetune the model was collected.
As this setup precludes evaluating models trained after 2020 on \dataset, we release our data collection pipeline
so that researchers can produce evaluation datasets from future versions of Wikipedia.

\section{Dataset Collection and Analysis}


As discussed in the introduction, keeping track of new information and then updating articles to reflect that information requires a massive amount of manual effort.
Thus, in order to scalably collect sufficient data for training and evaluating \taskacronym systems, some amount of automation is likely required.
In this section we introduce the \dataset dataset and associated data collection pipeline, 
which allows the automatic collection of high-quality training and evaluation data for \taskacronym from pairs of Wikipedia snapshots.


\begin{table}[!tb]
    \centering
    \small
    \begin{tabular}{p{5.5em}ccc}
        \toprule
                & \multirow{2}{*}{Train} & \multicolumn{2}{c}{Test} \\
                & & Silver & Gold \\
        \midrule
        Years & '19-'20 & '20-'21 & '20-'21 \\ 
        Articles & 114K & 54K & 914 \\
        Edits & 407K & 182K & 3.0K \\
        Subst. Edits & 135K & 62K & 1.3K \\
        Evidence & 720K & 315K & 7.7K \\
        Content Sel. & 93K & 42K & 913 \\
        \bottomrule
    \end{tabular}
    \caption{{\bf Dataset Statistics.} We use 10\% of the training data as our validation data.}
    \label{tab:dataset-stats}
\end{table}

\subsection{Pipeline}\label{section:pipeline}
Our data collection pipeline produces distantly annotated training and evaluation data from pairs of Wikipedia snapshots.
We will refer to the earlier snapshot as the \emph{source} snapshot, and the later snapshot as the \emph{target} snapshot.

\paragraph{Step 1. Collect Article Updates}
We compute the diff between the introductory sections of articles appearing in both the \emph{source} and \emph{target} snapshot to identify all of the material that has been updated (which will serve as $\src$ and $\tgt$). 
We also compute the diff between the non-introductory sections of articles to find new mentions of the subjects of other articles (which will serve as $\evidence$). 
These mentions can take the form of sentences in the text, as well as new table rows and list entries.
Entities are disambiguated using Wikipedia hyperlinks.

\paragraph{Step 2. Filter Stylistic Updates}
A large number of edits to Wikipedia are stylistic~\cite{daxenberger-gurevych-2012-corpus}, and are therefore irrelevant to our task.
In the next step of the pipeline, we attempt to filter articles that have only been superficially edited by keeping only those where at least one new \emph{added entity} appears in the \emph{target} snapshot.

\paragraph{Step 3. Identify Supporting Evidence}
In the last step of our pipeline, we seek to determine which pieces of evidence in $\evidence$ justify each of the updated sentences in $\tgt$. 
To do so, we make the following distant supervision assumption: an updated sentence $a \in \tgt$ containing an \emph{added entity} $s'$ is substantiated by a piece of evidence $E \in \evidence$ only if $s'$ is also mentioned in $E$.
The accuracy of the annotations produced by this assumption will be measured in Section~\ref{section:gold-data}.

\vspace{1em}
Our pipeline is implemented using Apache Beam,\footnote{\url{https://beam.apache.org/}} to allow for distributed processing.
We plan on releasing the code upon publication to enable other users to produce \taskacronym data from future Wikipedia snapshots, as well as languages other than English.

\subsection{\dataset}
We run our pipeline on English Wikipedia snapshots from Nov. 20, 2019 to Nov. 20, 2020 to produce the training dataset, and from Nov. 20, 2020 to June 1, 2021 to produce the evaluation dataset.
Detailed statistics are provided in Table~\ref{tab:dataset-stats}.
On average, there are around 3 to 4 updates per article, and around 7 pieces of associated evidence.
About 80\% of updates require some form of content selection, i.e., ignoring some evidence, when performing updates.

We find that only a third of the updates are substantiated by one or more pieces of evidence according to our distant supervision assumption. 
Thus, the remaining updates are either: a) superficial changes to the source article, or b) additions of new claims that are unsupported with respect to the collected evidence.
The latter is a particular issue as these claims can cause the model to learn to hallucinate during training, and should be impossible for the model to guess during evaluation.
Through the usage of human annotations and carefully selected evaluation metrics we will study the extent to which this is an issue throughout the rest of the paper.

\begin{figure}[!h]
    \centering
    \includegraphics[width=\linewidth]{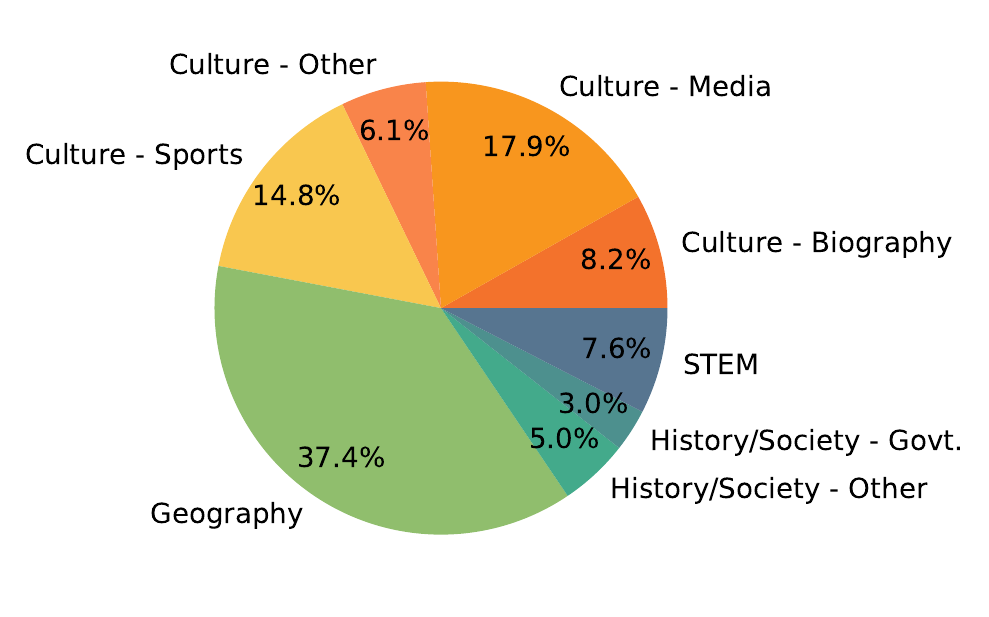}
    \caption{
        {\bf Topic Distribution.}
    }
    \label{fig:pie-chart}
\end{figure}

We categorize articles in our dataset using the Wikimedia Foundation's topic model~\cite{asthana2018wikitopic}.
The distribution of topics is displayed in Figure~\ref{fig:pie-chart}.
We find that the majority (approximately 50\%) of updates deal with cultural topics (e.g., sports, media, personal biographies), and geographic entities (e.g., countries, states) which intuitively are likely to be affected by current events.
while there are few updates to STEM- and history-related articles.

\begin{table}[!tb]
    \centering
    \small
    \begin{tabular}{ccccc}
        \toprule
        \multicolumn{3}{c}{\deltarouge} & \multicolumn{2}{c}{Entity} \\ 
        \cmidrule(lr){1-3} \cmidrule(lr){4-5} 
        1 & 2 & L & Prec. & Recall \\
        \midrule
         87.4 & 84.6 & 87.1 & 91.8 & 94.6 \\
        \bottomrule
    \end{tabular}
    \caption{
        {\bf Inter-Annotator Agreement}.
    }
    \label{tab:agreement}
\end{table}

\subsection{Gold Evaluation Data}
\label{section:gold-data}

To address the issue of unsupported claims during evaluation, we hired a team of 9 annotators to produce a \emph{``gold''} evaluation subset of our test dataset.
We collect annotations for 914 update-evidence pairs where
each instance is corrected to ensure that all of the updates are supported.
For the remainder of the paper we will refer to the distantly supervised test dataset annotations as \emph{``silver''}.

\paragraph{Annotation Process}
For each instance, annotators were shown the source article, evidence, and a marked up copy of the target article.
In the marked up article, each updated sentence was highlighted and prefixed with reference labels to the supporting evidence identified by our pipeline.
The correction process proceeded in two steps.
In the first step, annotators were asked to highlight all of the unsupported claims and incorrect reference labels in the target article.
In the second step, annotators were then asked to remove the unsupported text and minimally update the article to preserve fluency.

Annotators attended an initial 30 minute training and were provided regular feedback from the authors during the early stages of annotation.
To ensure data quality, an additional annotator was hired with the sole job of checking the other annotator's work and correcting their mistakes.
In total annotators spent roughly 500 hours on annotation.
The annotation interface and a completed annotation are shown in Figure~\ref{fig:annotator-interface} in the Appendix.

\paragraph{Agreement}
We measure annotator agreement using a subset of 100 instances that were annotated by multiple annotators.
Following \citet{Chen2015MicrosoftCC} and \citet{shi-etal-2021-descgen}, we quantify agreement by computing the evaluation metrics described in Section~\ref{section:evaluation}.
The results are provided in Table~\ref{tab:agreement}.
We observe high inter-annotator agreement with all scores in the 80s and 90s.

\paragraph{Analysis}
Statistics for the gold evaluation dataset are provided in Table~\ref{tab:dataset-stats}.
Overall, they closely resemble the statistics for the distantly supervised data with one exception: the fraction of substantiated updates has increased. 

To measure the quality of our silver data, we re-apply the approach used to measure inter-annotator agreement to compute agreement between the gold and silver annotations.
We also measure the \emph{reference agreement}, i.e., the fraction of reference labels kept by the annotators.
Results are provided in Table~\ref{tab:silver-gold}.
We find that agreement is high with most scores in the 80s, a strong indication that the data produced by our pipeline is high quality.
In particular, the high \deltarouge scores provide further evidence that only a small amount of the updated text in the weakly supervised data is unsupported, while the high reference agreement indicates that our distant supervision assumption is usually accurate.

\begin{table}[!tb]
    \centering
    \small
    \begin{tabular}{cccccc}
        \toprule
        \multicolumn{3}{c}{\deltarouge} & \multicolumn{2}{c}{Entity} & \multirowcell{2}{Reference\\ Agreement} \\
        \cmidrule(lr){1-3} \cmidrule(lr){4-5}
        1 & 2 & L & Prec. & Recall \\ 
        \midrule
        83.7 & 81.2 & 83.4 & 90.4 & 100.0 & 84.5 \\
        \bottomrule
    \end{tabular}
    \caption{
        {\bf Gold and Silver Annotation Agreement}.
        Quality of Silver Annotations by using the Gold as the reference.
    }
    \label{tab:silver-gold}
\end{table}

\begin{table*}[t]
\centering
\begin{subtable}[t]{0.66\textwidth}
    \small
    \centering
    \begin{tabular}{lcccccc}
        \toprule
        & \multicolumn{3}{c}{\deltarouge} & \multicolumn{2}{c}{Entity} & Unsup.
        \\
        \cmidrule(lr){2-4} \cmidrule(lr){5-6} \cmidrule(lr){7-7} 
        & 1 & 2 & L & Prec. & Recall &  Tokens \\
        \midrule
        Copy Source                       & 0.0 & 0.0 & 0.0 & 0.0 & 0.0 & 0.00 \\
        \hspace{0.2em} + All Evidence     & 18.8 & 6.9 & 12.0 & 37.9 & $64.9^*$ & 0.00 \\
        \midrule
        T5-Large                          & 31.1 & 18.4 & 24.4 & 52.7 & 44.9 & 2.67 \\
        \hspace{0.2em} + Evidence Input   & 44.3 & 29.4 & 36.8 & 62.2 & 50.7 & 2.34 \\
    
        \midrule
        \model-Small 
                                     & 41.2 & 27.3 & 35.3 & 62.4 & 44.9 & 1.71 \\
        \model-Base           & 47.0 & 32.1 & 39.7 & 62.2 & \bf 54.9 & 2.28 \\
        \model-Large          & 46.3 & 32.4 & 39.6 & 67.2 & 53.1 & \bf 1.54 \\
        \model-3B             & \bf 47.4 & \bf 34.0 & \bf 41.1 & \bf 69.9 & 52.5 & 1.58 \\
        \bottomrule
    \end{tabular} \\
    $^*$Entity recall is not 100\% for the Copy Source + All Evidence baseline due to lexical variation in entity mentions.
    \caption{}
    \label{tab:gold-baseline}
\end{subtable}%
\begin{subtable}[t]{0.32\textwidth}
        \centering
        \begin{tabular}{lc}
            \toprule
            {\bf Grounded Updates} & 50 \\
            \hspace{0.5em} Additional Content & 15 \\
            \hspace{0.5em} Missing Content & 22 \\
            \midrule
            {\bf Ungrounded Updates} & 35 \\
            \hspace{0.5em} Number/Date & 21 \\
            \hspace{0.5em} Distorted Evidence & 11 \\
            \hspace{0.5em} Hallucination & 14 \\
            \toprule
            {\bf No Updates} & 14 \\
            \bottomrule
        \end{tabular}
        \caption{}
        \label{tab:error-analysis}
\end{subtable}
\caption{
    {\bf (a)  Model Results on Gold Evaluation Data.}
    \model outperforms T5 models in all metrics.
    {\bf (b) Error Analysis for \model-3B}. We find that the model makes correct, grounded updates on 50\% of the inspected articles. For incorrect updates, ungrounded numbers/dates are one of the main sources of error.
}
\end{table*}

\section{Methods}

In this section we introduce baseline methods to establish initial benchmark results on \dataset.
We consider trivial approaches that copy task inputs, as well as T5, a neural sequence-to-sequence baseline which has shown strong performance on related tasks such as summarization~\cite{Raffel2020T5,rothe-etal-2021-thorough}
We additionally introduce \model, a variant of T5 that produces a sequence of edits instead of the entire updated text, and employs additional tweaks to improve performance.

\subsection{Copy Baselines}
The first set of baselines we introduce are trivial methods that merely copy the input.
We consider two variants:
\begin{itemize}[nosep,leftmargin=*]
\item {\bf Copy Source}: Generates a copy of the source article, and
\item {\bf Copy Source + Evidence}: Generates a copy of the source article concatenated with the evidence.
\end{itemize}
Our evaluation metrics only apply to unstructured text, however the evidence may contain structured tables.
In order to convert these tables to text, we apply a conventional linearization scheme~\cite{lebret-etal-2016-neural,wiseman-etal-2017-challenges} that separates table entries using row and column delimiters.

\subsection{T5}
\label{section:T5}
T5~\cite{Raffel2020T5} is a pretrained sequence-to-sequence~\cite{Sutskever2014SequenceTS} model based on the transformer architecture~\cite{Vaswani2017AttentionIA}.
Similar to the previous section we experiment with two variants:
\begin{itemize}[nosep,leftmargin=*]
\item {\bf T5}: Only includes the source article in its input,
\item {\bf T5 + Evidence Inputs}: Includes both the source article and evidence in the input.
\end{itemize}
Tabular inputs are linearized using the same approach described in the previous section.
Experiments are performed using the JAX-based T5X library.\footnote{https://github.com/google-research/t5x}
Hyperparameters and additional training details are described in Appendix~\ref{appendix:model}.

\subsection{\model}
\label{section:model}


Lastly, we introduce \model, which improves upon the T5-based approach described in the previous section through the usage of a compressed output format that removes the need to write the entire update from scratch and encourages content planning.
The output is modified in two ways:

First, as the majority of text in the target article is copied from the source, we replace any copied sentence with a single \emph{copy token} identifying the sentence, e.g., if the second sentence is copied it is replaced by the token \texttt{[2]}.
Similar to a copy mechanism~\cite{see-etal-2017-get}, this allows the model to dedicate less capacity to repeating sequences from the input.
As the resulting output resembles that produced by the \texttt{diff} data comparison utility, we refer to this as a diff-formatted output.

Second, before each update we insert a sequence of \emph{reference tokens} identifying the pieces of evidence that support the update, e.g., if the first and third piece of evidence in $\evidence$ support an update then the update is prefaced by \texttt{(1)(3)}.
This approach, inspired by the use of entity chains for summarization~\cite{Narayan2021PlanningWE}, trains the model to plan which references to use before generating an update.
These reference tokens are removed from the output text of the model prior to computing the evaluation metrics.

An example of the \model output format is provided in Figure~\ref{fig:edit5-format}, and a comparison to the T5 output format is provided in Appendix~\ref{appendix:output-formats}.
Training details and hyperparameters match the setup described in Section~\ref{section:T5}.

\begin{figure}[!t]
    \footnotesize
    \sf
    \begin{justify}
\nlp{\RaggedRight(2) Tom Krister Kristensson (born 30 April 1991) is a Swedish rally driver, who drives in the Junior World Championship.
    [1]
    [2]
    (1) In the 2019 season of JWRC, Tom finished second behind Jan Solans.
    (2) The next season he went on to become the 2020 Junior World Rally champion.}
    \caption{
        {\bf \model Output Format.} Instead of generating the fully updated text, \model generates sequences of edited sentences, copy tokens (e.g., {\tt [2]}, which means copy the second sentence), and reference tokens (e.g., {\tt (1)}, which means the following sentence should use the first piece of evidence).
    }
    \label{fig:edit5-format}
    \end{justify}
\end{figure}

\section{Results and Analysis}
\label{section:results-analysis}

Baseline results on the gold evaluation data are provided in Table~\ref{tab:gold-baseline}, and ablation results are provided in Appendix~\ref{section:ablation-study}.
In general, we find that the copy baselines perform worse than T5 and T5 performs worse than \model.
Notably, the copy source baseline rightfully scores zero on all metrics, while we will later find that it obtains a high ROUGE score.

Although our models are trained on silver data, they still obtain good performance on the gold evaluation set.
This shows the high quality of our silver data collection pipeline, and T5's ability to generate reasonable updates based on the evidence.

For the T5 baselines, we find that adding evidence to the input results significant increase in all metrics, demonstrating that using the evidence is crucial to obtaining good performance.

\model obtains additional 3-5\% absolute increase in all performance metrics compared to T5, establishing \model as a strong baseline for future systems to be compared against.
The reduction of unsupported entity tokens implies that \model hallucinates less frequently than T5 models. Results are provided for different model sizes to illustrate how performance scales with parameter counts.


\begin{figure*}[!t]
    \centering
    \includegraphics{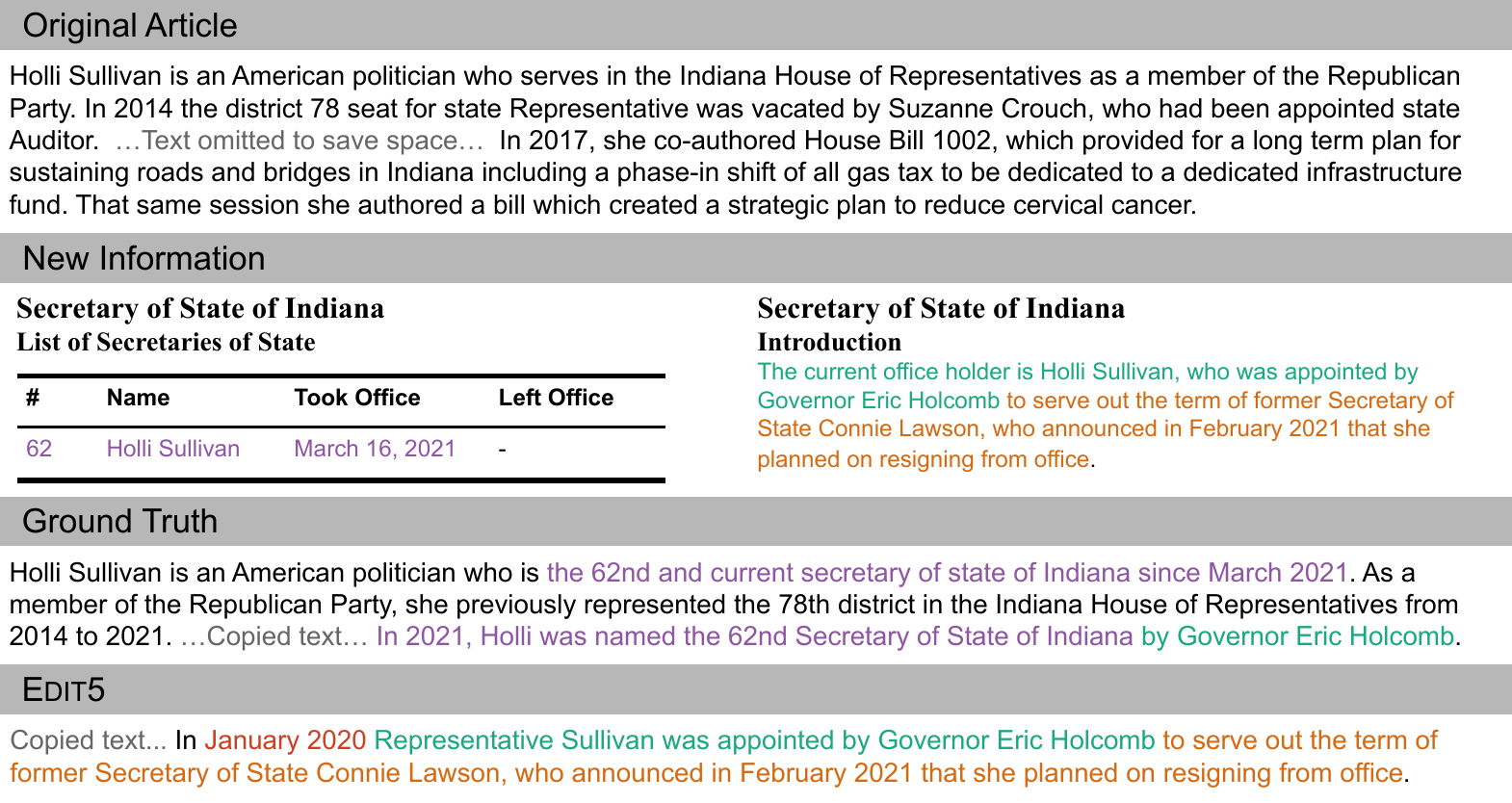}
    \caption{
        {\bf Example \model Output vs Ground Truth.}
        Color coding indicates alignment between the new information and the edits.
        \model updates the original article by paraphrasing sentences from the textual evidence, however misses relevant information in the table, and generates an incorrect date.
    }
    \label{fig:qualitative-results}
\end{figure*}


\paragraph{Example Output}
An example \model output is provided in Figure~\ref{fig:qualitative-results}, and additional outputs in Appendix~\ref{appendix:examples}.
The examples illustrate important features of the task.
In Figure~\ref{fig:qualitative-results} the goal is to update the Wikipedia article for Holli Sullivan to reflect her new role of Secretary of State of Indiana.
In the reference, this information is reflected in an updated version of the first sentence as well as in a newly added last sentence.
An additional sentence is added after the first sentence paraphrasing the introduction of the source article, which describes Sullivan's previous position as a member of the Indiana House of Representatives.

In the \model output for this example, information is only added at the end of the article.
While the model correctly states that Sullivan was appointed to be Secretary of State by Governor Eric Holcomb, as well as includes additional context surrounding Sullivan's appointment that is paraphrased from the evidence, there are some issues with the output. 
First, because the first sentence of the article is not updated there is conflicting information about Sullivan's current position.
Second, the added sentence hallucinates that Sullivan was appointed in January 2020 when she was actually appointed in March 2021, a fact that directly appears in the evidence.

\paragraph{Categorizing Errors}
To better understand the types of errors made by \model, we review a random sample of 100 of its predictions on the gold evaluation data and categorize them as either: \emph{grounded updates}, meaning all generated claims are  supported, \emph{ungrounded updates}, meaning at least one unsupported claim appears in the output, or \emph{no updates}, meaning the model did not predict any updates.
For grounded updates we additionally keep track of how many updates include \emph{additional content} not present in the ground truth update, or are \emph{missing content} that appears in the ground truth update.
For ungrounded updates we track whether an incorrect \emph{number/date} appears in the update, the model \emph{distorted evidence}, i.e., paraphrased or combined claims in the evidence in a way that changed their meaning, or \emph{hallucinated} new claims unrelated to the evidence.

The results of this analysis are presented in Table~\ref{tab:error-analysis}.
We find that \model makes no mistakes on half of the examples, however a substantial portion of these updates had some issue with content selection.
Of the incorrect updates, the most common mistake was incorrect numbers and dates, followed by hallucinations, and finally distorted evidence.
This suggests that improving numeracy could be a fruitful line of study in future work on this task.


\begin{table}[!t]
\small
\centering
 \begin{tabular}{lccc}
        \toprule
        & \multicolumn{3}{c}{ROUGE} \\
        \cmidrule(lr){2-4}
        &1 & 2 & L \\
        \midrule
   Copy Source &        78.1 & 69.3 & 75.0 \\
                               
  T5-Large & 57.0 & 44.2 & 49.5 \\
  \model-Large   &    78.6 & 69.1 & 72.7 \\
        \bottomrule
    \end{tabular}
    \caption{
        {\bf ROUGE Scores Are Insensitive to Edits.}
    }
    \label{tab:busted-rouge}
\end{table}
\paragraph{ROUGE is Problematic} 
We provide ROUGE F-scores for each of the baseline models on the gold evaluation data in Table~\ref{tab:busted-rouge}.
In contrast to the previous results, we find that the simple copy source baseline attains a strong score of 77.4 despite making no updates.
This is better than the T5 baseline results and comparable to the \model results.
This illustrates the importance of evaluating on updates rather than the whole text.

\paragraph{Silver Data is Useful for Evaluation}
The results in Section~\ref{section:gold-data} demonstrate high agreement between the silver and gold evaluation data which begs the question: can silver data be used in place of gold data for evaluation?
To answer this, we measure the Spearman rank correlation between the gold baseline results in Table~\ref{tab:gold-baseline} and silver baseline results (provided in Table~\ref{tab:silver-baseline} of the Appendix to save space).
Rank correlations for each of the metrics are shown in Table~\ref{tab:spearman}.
Overall we find high rank correlation for each of the metrics, which suggests silver evaluation performance is a reliable indicator of gold performance.
Thus, models whose pretraining data overlaps \dataset may be evaluated and compared on data produced by running our pipeline on future Wikipedia snapshots without requiring further human evaluation.

\begin{table}[!t]
    \centering
    \small
    \begin{tabular}{cccccc}
        \toprule
         \multicolumn{3}{c}{\deltarouge} & \multicolumn{2}{c}{Entity} & Unsup.\\
        \cmidrule(lr){1-3} \cmidrule(lr){4-5} \cmidrule(lr){6-6}
         1 & 2 & L & Prec. & Rec. & tokens \\
        \midrule
        100.0 & 100.0 & 94.3 & 75.4 & 92.8 & 92.8 \\
        \bottomrule
    \end{tabular}
    \caption{
        {\bf Spearman Rank Correlation Between Gold and Silver Performance Metrics}.
    }
    \label{tab:spearman}
\end{table}

\paragraph{Controllability}
The improvement we obtained from
\model over T5 implies that more controls can be added into the model.
In this section we investigate whether additional control provided by the users can improve the overall generations.
We follow \citet{Keskar2019CTRLAC} and \citet{Narayan2021PlanningWE}, and provide more detailed instruction by adding \emph{control codes}, i.e., special tokens, to the {\em input} that instruct the model whether to add, copy, edit or remove a sentence, as well as which evidence to use when making an addition or edit.
We use the target text to provide oracle labels for the control code, and see if the \model can take advantage of the codes.
Example inputs and predictions are provided in Figure~\ref{fig:controllable} of the Appendix.

Results on the gold evaluation data are provided in Table~\ref{tab:controllability}.
Including oracle control codes in the input produces a substantial 10\% absolute improvement in all metrics besides unsupported tokens.
This demonstrates that increased user control has the potential to produce updates that more closely resemble the desired output.

\begin{table}[!tb]
    \centering
    \small
    \begin{tabular}{lcccccc}
        \toprule
        & \multicolumn{3}{c}{\deltarouge} & \multicolumn{2}{c}{Entity} & Unsup.  \\
        \cmidrule(lr){2-4} \cmidrule(lr){5-6} \cmidrule(lr){7-7}
        & 1 & 2 & L & Prec. & Rec. & Tokens \\
        \midrule
        \model      & 46.3 & 32.4 & 39.6 & 67.2 & 53.1 & \bf 1.54\\
        Control     & \bf 57.6 & \bf 42.1 & \bf 50.2 & \bf 70.5 & \bf 64.5 & 2.42 \\
        \bottomrule
    \end{tabular}
    \caption{
        {\bf Controllability}. Using control codes that indicate which sentences to delete, add or edit, and which evidence to use, can greatly improve generation.
    }
    \label{tab:controllability}
\end{table}

\section{Related Work}

Early work on writing assistants largely focuses on grammar error correction; for a survey see \citet{Wang2020ACS}.
Neural models have expanded the capabilities of writing assistants to solve a wider variety of tasks including: autocompletion~\cite{chen2019gmail}, and following rhetorical directives such as paraphrasing, elaborating, etc.~\cite{sun-etal-2021-iga}.
In this work, we seek to expand these capabilities further to producing grounded updates, which has been previously studied by \citet{kang-etal-2019-pomo}, however only for post-modifier generation.

As our primary focus is on writing grounded updates to Wikipedia articles, our work is closely related to existing works on Wikipedia article generation, which generally uses one of two approaches: data-to-text generation~\cite{lebret-etal-2016-neural,Bao2018TabletoTextDT,parikh-etal-2020-totto,chen-etal-2021-wikitablet,cheng-etal-2020-ent}, or multi-document summarization~\cite{Banerjee2016WikiWriteGW,Liu2018GeneratingWB,shi-etal-2021-descgen}.
In particular, the hyperlink-based approach for associating evidence to articles is directly inspired by these works, and our annotation procedure for removing unsupported text directly draws from \citet{parikh-etal-2020-totto}.


Determining which facts contradict claims in the existing article is a central topic of work on fact extraction and verification~\cite{thorne-etal-2018-fever}.
Recently, \citet{schuster-etal-2021-get} introduced the \textsc{Vitamin-C} dataset of factual revisions to Wikipedia articles and the task of factually consistent generation. This work differs from \taskacronym in that it only focuses on sentences and does not require adding new facts or content selection.

Our work is also related to the TAC 2008 Update Summarization Task~\cite{Dang2008OverviewOT},
which
involves summarizing information about a topic that does not overlap with an existing summary,
instead of updating an existing summary to reflect new information.

\section{Conclusion and Future Work}

In this work we introduced \taskacronym, a novel text generation task where the goal is to update an article to reflect new information about its subject.
To enable research on this task, we formulated a pipeline for extracting weakly supervised training and evaluation data from pairs of Wikipedia snapshots, and collected data for the years 2019-2020 and 2020-2021, as well as human annotated gold evaluation data.
We additionally provided results for several strong baselines, that demonstrate both the feasibility of this task, as well as strong correlation between gold and distantly supervised data evaluation performance that establishes the trustworthiness of future data produced using our pipeline.

This work lays the foundation for future research into making faithful updates to entries in textual knowledge bases.
One limitation of this work is that the metrics we use to evaluate faithfulness are all entity-centric, and thus may overstate the performance of models whose edits include the correct entities but misspecify the relations between them or other facets of the evidence.
Accordingly, one promising direction for future work on this task is to develop more robust metrics for measuring faithfulness. 

An additional promising direction for future work is to consider open settings of evidence collection, where other forms of updated information such as excerpts from news articles could be used to justify edits in place of updates to other entries in the same knowledge base.
Relatedly, we also recommend studying this task in streaming settings, where updates arrive in sequential fashion, in addition to the batch setting considered in this work.

Finally, there are a number of promising directions for improving model performance on this task.
In particular, copy mechanisms~\cite{see-etal-2017-get} have been widely used in data-to-text tasks~\cite{wiseman-etal-2017-challenges}, and may help mitigate issues such as the mistranscribed dates we saw in Section~\ref{section:results-analysis}.



\section*{Acknowledgements}
We would like to thank Kelvin Guu, Kenton Lee, Kristina Toutanova, and the anonymous reviewers for their helpful feedback.
This work is funded in part by the DARPA MCS program under Contract No. N660011924033. 
Additionally, Robert is supported in part by the Irvine Initiative in AI, Law, and Society fellowship.

\section*{Ethical Considerations}

This paper introduces a dataset and system for updating an existing piece of text to incorporate information from external evidence.
Depending on the veracity of the external evidence, systems for solving this task could potentially be abused by bad actors to spread misinformation.

\clearpage

\bibliography{references-rebib}

\begin{thebibliography}{32}
\expandafter\ifx\csname natexlab\endcsname\relax\def\natexlab#1{#1}\fi

\bibitem[{Asthana and Halfaker(2018)}]{asthana2018wikitopic}
Sumit Asthana and Aaron Halfaker. 2018.
\newblock \href {https://doi.org/10.1145/3274290} {With few eyes, all hoaxes
  are deep}.
\newblock \emph{Proc. ACM Hum.-Comput. Interact.}, 2(CSCW).

\bibitem[{Banerjee and Mitra(2016)}]{Banerjee2016WikiWriteGW}
Siddhartha Banerjee and Prasenjit Mitra. 2016.
\newblock \href {http://www.ijcai.org/Abstract/16/389} {Wikiwrite: Generating
  wikipedia articles automatically}.
\newblock In \emph{Proceedings of the Twenty-Fifth International Joint
  Conference on Artificial Intelligence, {IJCAI} 2016, New York, NY, USA, 9-15
  July 2016}, pages 2740--2746. {IJCAI/AAAI} Press.

\bibitem[{Bao et~al.(2018)Bao, Tang, Duan, Yan, Lv, Zhou, and
  Zhao}]{Bao2018TabletoTextDT}
Junwei Bao, Duyu Tang, Nan Duan, Zhao Yan, Yuanhua Lv, Ming Zhou, and Tiejun
  Zhao. 2018.
\newblock \href
  {https://www.aaai.org/ocs/index.php/AAAI/AAAI18/paper/view/16138}
  {Table-to-text: Describing table region with natural language}.
\newblock In \emph{Proceedings of the Thirty-Second {AAAI} Conference on
  Artificial Intelligence, (AAAI-18), the 30th innovative Applications of
  Artificial Intelligence (IAAI-18), and the 8th {AAAI} Symposium on
  Educational Advances in Artificial Intelligence (EAAI-18), New Orleans,
  Louisiana, USA, February 2-7, 2018}, pages 5020--5027. {AAAI} Press.

\bibitem[{Chen et~al.(2019)Chen, Lee, Bansal, Cao, Zhang, Lu, Tsay, Wang, Dai,
  Chen, Sohn, and Wu}]{chen2019gmail}
Mia~Xu Chen, Benjamin~N. Lee, Gagan Bansal, Yuan Cao, Shuyuan Zhang, Justin Lu,
  Jackie Tsay, Yinan Wang, Andrew~M. Dai, Zhifeng Chen, Timothy Sohn, and
  Yonghui Wu. 2019.
\newblock \href {https://doi.org/10.1145/3292500.3330723} {Gmail smart compose:
  Real-time assisted writing}.
\newblock In \emph{Proceedings of the 25th {ACM} {SIGKDD} International
  Conference on Knowledge Discovery {\&} Data Mining, {KDD} 2019, Anchorage,
  AK, USA, August 4-8, 2019}, pages 2287--2295. {ACM}.

\bibitem[{Chen et~al.(2021)Chen, Wiseman, and
  Gimpel}]{chen-etal-2021-wikitablet}
Mingda Chen, Sam Wiseman, and Kevin Gimpel. 2021.
\newblock \href {https://doi.org/10.18653/v1/2021.findings-acl.17}
  {{W}iki{T}able{T}: A large-scale data-to-text dataset for generating
  {W}ikipedia article sections}.
\newblock In \emph{Findings of the Association for Computational Linguistics:
  ACL-IJCNLP 2021}, pages 193--209, Online. Association for Computational
  Linguistics.

\bibitem[{Chen et~al.(2015)Chen, Fang, Lin, Vedantam, Gupta, Doll{\'a}r, and
  Zitnick}]{Chen2015MicrosoftCC}
Xinlei Chen, Hao Fang, Tsung-Yi Lin, Ramakrishna Vedantam, Saurabh Gupta, Piotr
  Doll{\'a}r, and C.~Lawrence Zitnick. 2015.
\newblock Microsoft coco captions: Data collection and evaluation server.
\newblock \emph{ArXiv}, abs/1504.00325.

\bibitem[{Cheng et~al.(2020)Cheng, Wu, Bing, Zhang, Jie, Lu, and
  Si}]{cheng-etal-2020-ent}
Liying Cheng, Dekun Wu, Lidong Bing, Yan Zhang, Zhanming Jie, Wei Lu, and Luo
  Si. 2020.
\newblock \href {https://doi.org/10.18653/v1/2020.emnlp-main.90} {{ENT}-{DESC}:
  Entity description generation by exploring knowledge graph}.
\newblock In \emph{Proceedings of the 2020 Conference on Empirical Methods in
  Natural Language Processing (EMNLP)}, pages 1187--1197, Online. Association
  for Computational Linguistics.

\bibitem[{Dang and Owczarzak(2008)}]{Dang2008OverviewOT}
Hoa~Trang Dang and Karolina Owczarzak. 2008.
\newblock Overview of the tac 2008 update summarization task.
\newblock \emph{Theory and Applications of Categories}.

\bibitem[{Daxenberger and Gurevych(2012)}]{daxenberger-gurevych-2012-corpus}
Johannes Daxenberger and Iryna Gurevych. 2012.
\newblock \href {https://aclanthology.org/C12-1044} {A corpus-based study of
  edit categories in featured and non-featured {W}ikipedia articles}.
\newblock In \emph{Proceedings of {COLING} 2012}, pages 711--726, Mumbai,
  India. The COLING 2012 Organizing Committee.

\bibitem[{Guu et~al.(2020)Guu, Lee, Tung, Pasupat, and Chang}]{guu2020realm}
Kelvin Guu, Kenton Lee, Zora Tung, Panupong Pasupat, and Ming-Wei Chang. 2020.
\newblock \href {http://arxiv.org/abs/2002.08909} {Realm: Retrieval-augmented
  language model pre-training}.

\bibitem[{Kang et~al.(2019)Kang, Logan, Chu, Chen, Dua, Gimpel, Singh, and
  Balasubramanian}]{kang-etal-2019-pomo}
Jun~Seok Kang, Robert Logan, Zewei Chu, Yang Chen, Dheeru Dua, Kevin Gimpel,
  Sameer Singh, and Niranjan Balasubramanian. 2019.
\newblock \href {https://doi.org/10.18653/v1/N19-1089} {{P}o{M}o: Generating
  entity-specific post-modifiers in context}.
\newblock In \emph{Proceedings of the 2019 Conference of the North {A}merican
  Chapter of the Association for Computational Linguistics: Human Language
  Technologies, Volume 1 (Long and Short Papers)}, pages 826--838, Minneapolis,
  Minnesota. Association for Computational Linguistics.

\bibitem[{Keskar et~al.(2019)Keskar, McCann, Varshney, Xiong, and
  Socher}]{Keskar2019CTRLAC}
Nitish~Shirish Keskar, Bryan McCann, Lav~R. Varshney, Caiming Xiong, and
  Richard Socher. 2019.
\newblock Ctrl: A conditional transformer language model for controllable
  generation.
\newblock \emph{ArXiv}, abs/1909.05858.

\bibitem[{Krishna et~al.(2021)Krishna, Roy, and
  Iyyer}]{krishna-etal-2021-hurdles}
Kalpesh Krishna, Aurko Roy, and Mohit Iyyer. 2021.
\newblock \href {https://doi.org/10.18653/v1/2021.naacl-main.393} {Hurdles to
  progress in long-form question answering}.
\newblock In \emph{Proceedings of the 2021 Conference of the North American
  Chapter of the Association for Computational Linguistics: Human Language
  Technologies}, pages 4940--4957, Online. Association for Computational
  Linguistics.

\bibitem[{Lebret et~al.(2016)Lebret, Grangier, and
  Auli}]{lebret-etal-2016-neural}
R{\'e}mi Lebret, David Grangier, and Michael Auli. 2016.
\newblock \href {https://doi.org/10.18653/v1/D16-1128} {Neural text generation
  from structured data with application to the biography domain}.
\newblock In \emph{Proceedings of the 2016 Conference on Empirical Methods in
  Natural Language Processing}, pages 1203--1213, Austin, Texas. Association
  for Computational Linguistics.

\bibitem[{Lin(2004)}]{Lin2004ROUGEAP}
Chin-Yew Lin. 2004.
\newblock \href {https://aclanthology.org/W04-1013} {{ROUGE}: A package for
  automatic evaluation of summaries}.
\newblock In \emph{Text Summarization Branches Out}, pages 74--81, Barcelona,
  Spain. Association for Computational Linguistics.

\bibitem[{Liu et~al.(2018)Liu, Saleh, Pot, Goodrich, Sepassi, Kaiser, and
  Shazeer}]{Liu2018GeneratingWB}
Peter~J. Liu, Mohammad Saleh, Etienne Pot, Ben Goodrich, Ryan Sepassi, Lukasz
  Kaiser, and Noam Shazeer. 2018.
\newblock \href {https://openreview.net/forum?id=Hyg0vbWC-} {Generating
  wikipedia by summarizing long sequences}.
\newblock In \emph{6th International Conference on Learning Representations,
  {ICLR} 2018, Vancouver, BC, Canada, April 30 - May 3, 2018, Conference Track
  Proceedings}. OpenReview.net.

\bibitem[{Longpre et~al.(2021)Longpre, Perisetla, Chen, Ramesh, DuBois, and
  Singh}]{longpre-etal-2021-entity}
Shayne Longpre, Kartik Perisetla, Anthony Chen, Nikhil Ramesh, Chris DuBois,
  and Sameer Singh. 2021.
\newblock \href {https://doi.org/10.18653/v1/2021.emnlp-main.565} {Entity-based
  knowledge conflicts in question answering}.
\newblock In \emph{Proceedings of the 2021 Conference on Empirical Methods in
  Natural Language Processing}, pages 7052--7063, Online and Punta Cana,
  Dominican Republic. Association for Computational Linguistics.

\bibitem[{Narayan et~al.(2021)Narayan, Zhao, Maynez, Simoes, and
  McDonald}]{Narayan2021PlanningWE}
Shashi Narayan, Yao Zhao, Joshua Maynez, Gonccalo Simoes, and Ryan~T. McDonald.
  2021.
\newblock Planning with entity chains for abstractive summarization.
\newblock \emph{ArXiv}, abs/2104.07606.

\bibitem[{Parikh et~al.(2020)Parikh, Wang, Gehrmann, Faruqui, Dhingra, Yang,
  and Das}]{parikh-etal-2020-totto}
Ankur Parikh, Xuezhi Wang, Sebastian Gehrmann, Manaal Faruqui, Bhuwan Dhingra,
  Diyi Yang, and Dipanjan Das. 2020.
\newblock \href {https://doi.org/10.18653/v1/2020.emnlp-main.89} {{ToTTo}: A
  controlled table-to-text generation dataset}.
\newblock In \emph{Proceedings of the 2020 Conference on Empirical Methods in
  Natural Language Processing (EMNLP)}, pages 1173--1186, Online. Association
  for Computational Linguistics.

\bibitem[{Raffel et~al.(2020)Raffel, Shazeer, Roberts, Lee, Narang, Matena,
  Zhou, Li, and Liu}]{Raffel2020T5}
Colin Raffel, Noam Shazeer, Adam Roberts, Katherine Lee, Sharan Narang, Michael
  Matena, Yanqi Zhou, Wei Li, and Peter~J. Liu. 2020.
\newblock \href {http://jmlr.org/papers/v21/20-074.html} {Exploring the limits
  of transfer learning with a unified text-to-text transformer}.
\newblock \emph{Journal of Machine Learning Research}, 21(140):1--67.

\bibitem[{Rothe et~al.(2021)Rothe, Maynez, and
  Narayan}]{rothe-etal-2021-thorough}
Sascha Rothe, Joshua Maynez, and Shashi Narayan. 2021.
\newblock \href {https://doi.org/10.18653/v1/2021.emnlp-main.12} {A thorough
  evaluation of task-specific pretraining for summarization}.
\newblock In \emph{Proceedings of the 2021 Conference on Empirical Methods in
  Natural Language Processing}, pages 140--145, Online and Punta Cana,
  Dominican Republic. Association for Computational Linguistics.

\bibitem[{Schuster et~al.(2021)Schuster, Fisch, and
  Barzilay}]{schuster-etal-2021-get}
Tal Schuster, Adam Fisch, and Regina Barzilay. 2021.
\newblock \href {https://doi.org/10.18653/v1/2021.naacl-main.52} {Get your
  vitamin {C}! robust fact verification with contrastive evidence}.
\newblock In \emph{Proceedings of the 2021 Conference of the North American
  Chapter of the Association for Computational Linguistics: Human Language
  Technologies}, pages 624--643, Online. Association for Computational
  Linguistics.

\bibitem[{See et~al.(2017)See, Liu, and Manning}]{see-etal-2017-get}
Abigail See, Peter~J. Liu, and Christopher~D. Manning. 2017.
\newblock \href {https://doi.org/10.18653/v1/P17-1099} {Get to the point:
  Summarization with pointer-generator networks}.
\newblock In \emph{Proceedings of the 55th Annual Meeting of the Association
  for Computational Linguistics (Volume 1: Long Papers)}, pages 1073--1083,
  Vancouver, Canada. Association for Computational Linguistics.

\bibitem[{Shazeer and Stern(2018)}]{shazeer2018ada}
Noam Shazeer and Mitchell Stern. 2018.
\newblock \href {http://proceedings.mlr.press/v80/shazeer18a.html} {Adafactor:
  Adaptive learning rates with sublinear memory cost}.
\newblock In \emph{Proceedings of the 35th International Conference on Machine
  Learning, {ICML} 2018, Stockholmsm{\"{a}}ssan, Stockholm, Sweden, July 10-15,
  2018}, volume~80 of \emph{Proceedings of Machine Learning Research}, pages
  4603--4611. {PMLR}.

\bibitem[{Shi et~al.(2021)Shi, Joshi, and Zettlemoyer}]{shi-etal-2021-descgen}
Weijia Shi, Mandar Joshi, and Luke Zettlemoyer. 2021.
\newblock \href {https://doi.org/10.18653/v1/2021.acl-long.35} {{DESCGEN}: A
  distantly supervised datasetfor generating entity descriptions}.
\newblock In \emph{Proceedings of the 59th Annual Meeting of the Association
  for Computational Linguistics and the 11th International Joint Conference on
  Natural Language Processing (Volume 1: Long Papers)}, pages 415--427, Online.
  Association for Computational Linguistics.

\bibitem[{Sun et~al.(2021)Sun, Zhao, Manjunatha, Jain, Morariu, Dernoncourt,
  Srinivasan, and Iyyer}]{sun-etal-2021-iga}
Simeng Sun, Wenlong Zhao, Varun Manjunatha, Rajiv Jain, Vlad Morariu, Franck
  Dernoncourt, Balaji~Vasan Srinivasan, and Mohit Iyyer. 2021.
\newblock \href {https://doi.org/10.18653/v1/2021.emnlp-main.483} {{IGA}: An
  intent-guided authoring assistant}.
\newblock In \emph{Proceedings of the 2021 Conference on Empirical Methods in
  Natural Language Processing}, pages 5972--5985, Online and Punta Cana,
  Dominican Republic. Association for Computational Linguistics.

\bibitem[{Sutskever et~al.(2014)Sutskever, Vinyals, and
  Le}]{Sutskever2014SequenceTS}
Ilya Sutskever, Oriol Vinyals, and Quoc~V. Le. 2014.
\newblock \href
  {https://proceedings.neurips.cc/paper/2014/hash/a14ac55a4f27472c5d894ec1c3c743d2-Abstract.html}
  {Sequence to sequence learning with neural networks}.
\newblock In \emph{Advances in Neural Information Processing Systems 27: Annual
  Conference on Neural Information Processing Systems 2014, December 8-13 2014,
  Montreal, Quebec, Canada}, pages 3104--3112.

\bibitem[{Thorne et~al.(2018)Thorne, Vlachos, Christodoulopoulos, and
  Mittal}]{thorne-etal-2018-fever}
James Thorne, Andreas Vlachos, Christos Christodoulopoulos, and Arpit Mittal.
  2018.
\newblock \href {https://doi.org/10.18653/v1/N18-1074} {{FEVER}: a large-scale
  dataset for fact extraction and {VER}ification}.
\newblock In \emph{Proceedings of the 2018 Conference of the North {A}merican
  Chapter of the Association for Computational Linguistics: Human Language
  Technologies, Volume 1 (Long Papers)}, pages 809--819, New Orleans,
  Louisiana. Association for Computational Linguistics.

\bibitem[{Vaswani et~al.(2017)Vaswani, Shazeer, Parmar, Uszkoreit, Jones,
  Gomez, Kaiser, and Polosukhin}]{Vaswani2017AttentionIA}
Ashish Vaswani, Noam Shazeer, Niki Parmar, Jakob Uszkoreit, Llion Jones,
  Aidan~N. Gomez, Lukasz Kaiser, and Illia Polosukhin. 2017.
\newblock \href
  {https://proceedings.neurips.cc/paper/2017/hash/3f5ee243547dee91fbd053c1c4a845aa-Abstract.html}
  {Attention is all you need}.
\newblock In \emph{Advances in Neural Information Processing Systems 30: Annual
  Conference on Neural Information Processing Systems 2017, December 4-9, 2017,
  Long Beach, CA, {USA}}, pages 5998--6008.

\bibitem[{Wang et~al.(2020)Wang, Wang, Liu, and Liu}]{Wang2020ACS}
Yu~Wang, Yuelin Wang, Jie Liu, and Zhuo Liu. 2020.
\newblock A comprehensive survey of grammar error correction.
\newblock \emph{ArXiv}, abs/2005.06600.

\bibitem[{Wiseman et~al.(2017)Wiseman, Shieber, and
  Rush}]{wiseman-etal-2017-challenges}
Sam Wiseman, Stuart Shieber, and Alexander Rush. 2017.
\newblock \href {https://doi.org/10.18653/v1/D17-1239} {Challenges in
  data-to-document generation}.
\newblock In \emph{Proceedings of the 2017 Conference on Empirical Methods in
  Natural Language Processing}, pages 2253--2263, Copenhagen, Denmark.
  Association for Computational Linguistics.

\bibitem[{Çelikyilmaz et~al.(2020)Çelikyilmaz, Clark, and
  Gao}]{elikyilmaz2020EvaluationOT}
Asli Çelikyilmaz, Elizabeth Clark, and Jianfeng Gao. 2020.
\newblock Evaluation of text generation: A survey.
\newblock \emph{ArXiv}, abs/2006.14799.

\end{thebibliography}
\bibliographystyle{acl_natbib}

\newpage
\appendix

\renewcommand{\thefigure}{\appendixprefix\arabic{figure}}
\setcounter{figure}{0}
\renewcommand{\thetable}{\appendixprefix\arabic{table}} 
\setcounter{table}{0}
\renewcommand{\theequation}{\appendixprefix\arabic{equation}} 
\setcounter{equation}{0}




\clearpage
\section*{Appendix}


\section{Ablation Study}
\label{section:ablation-study}
We perform an ablation study to measure the impact of the modifications made to the target output of \model.
The results are provided in Table~\ref{tab:ablations}
We observe that both the diff format and including reference tokens have a positive impact on the evaluation metrics, with reference tokens having the larger impact.
 
\begin{table}[!h]
    \centering
    \small
    \hspace{-.2cm}
    \begin{tabular}{lcccccc}
        \toprule
        & \multicolumn{3}{c}{\deltarouge} & \multicolumn{2}{c}{Entity} & Unsupp.\\
        \cmidrule(lr){2-4} \cmidrule(lr){5-6} \cmidrule(lr){7-7}
        & 1 & 2 & L & Prec. & Rec. & Tokens \\
        \midrule
        \model                & \bf 46.3 & \bf 32.4 & \bf 39.6 & \bf 67.2 & \bf 53.1 & \bf 1.54 \\
        \hspace{0.1em} - Diff & 45.5 & 31.7 & 39.1 & 66.8 & 50.8 & 1.66 \\
        \hspace{0.1em} - Ref. & 45.1 & 31.6 & 38.8 & 66.3 & 50.7 & 1.89 \\
        \bottomrule
    \end{tabular}
    \caption{
        {\bf \model Ablations}.
    }
    \label{tab:ablations}
\end{table}


\section{Model Training Details}
\label{appendix:model}
Optimizer: AdaFactor~\cite{shazeer2018ada}, Batch Size: 128, Learning Rate: 1e-3, Dropout Rate: 0.1, Training Iterations: 30,000.
Training performed on a cluster of 16 2nd generation TPUs for <3B param models, and 32 TPUS for 3B parameter models.

\section{Silver Baseline Results}

\begin{table}[!h]
    \centering
    \small
    \begin{tabular}{lcccccc}
        \toprule
        & \multicolumn{3}{c}{\deltarouge} & \multicolumn{2}{c}{Target Entity} & Evid.
        \\
        \cmidrule(lr){2-4} \cmidrule(lr){5-6} \cmidrule(lr){7-7}
        & 1 & 2 & L & P & R & Acc \\
        \midrule
        T5-Large               & 26.8 & 15.9 & 22.3 & 56.3 & 29.8 & 2.33 \\
        \hspace{0.2em} + Evid. & 39.2 & 27.3 & 34.2 & 66.9 & 42.4 & 1.63 \\
        \midrule
        \model \\
        \hspace{0.2em}Small & 37.8 & 24.9 & 32.6 & 61.4 & 41.2 & 1.53 \\
        \hspace{0.2em}Base  & 42.8 & 28.7 & 36.4 & 60.5 & \bf 49.2 & 2.32 \\
        \hspace{0.2em}Large & 42.7 & 29.9 & 37.2 & 66.1 & 47.5 & 1.47 \\
        \hspace{0.2em}3B    & \bf 43.8 & \bf 31.5 & \bf 38.6 & \bf 68.4 & 48.6  & 1.53 \\
        \bottomrule
    \end{tabular}
    \caption{
        {\bf Baseline Results on Silver Evaluation Data.}
    }
    \label{tab:silver-baseline}
\end{table}

\onecolumn
\section{Input and Output Formats}
\label{appendix:output-formats}

\begin{figure}[!h]
    \small
    \sf
    \begin{justify}
\nlp{\RaggedRight(2) 
[0] Elizabeth Lynne Cheney (; born July 28, 1966) is an American attorney and politician serving as the U.S. Representative for since 2017. [1] Cheney is the House Republican Conference Chair, the third-highest position in GOP House leadership. [2] She is the third woman elected to that position after Deborah Pryce and Cathy McMorris Rodgers. [3] Cheney is the elder daughter of former Vice President Dick Cheney and Lynne Cheney. [4] She held several positions in the U.S. State Department during the George W. Bush administration. [5] She has been politically active on behalf of the Republican Party and is a co-founder of Keep America Safe, a nonprofit organization concerned with national security issues. [6] She was a candidate for the 2014 election to the United States Senate in Wyoming, challenging the three-term incumbent Mike Enzi, before withdrawing from the race. [7] In the House of Representatives, she holds the seat that was held by her father from 1979 to 1989. [8] She is known for her hawkish foreign policy views. [CONTEXT] (0) Andy Biggs U.S. House of Representatives - Tenure - 2021 storming of the United States Capitol On January 12, 2021, Biggs called on fellow GOP Representative Liz Cheney (R-WY) to resign from her leadership position within the Republican Caucus, after she voted in favor of Donald Trump's second impeachment. (1) 116th United States Congress Leadership - House of Representatives - Minority (Republican) leadership * House Minority Leader and Chair of the House Republican Steering Committee: Kevin McCarthy * House Minority Whip: Steve Scalise * Chair of the House Republican Conference: Liz Cheney * Vice Chair of the House Republican Conference: Mark Walker * Secretary of the House Republican Conference: Jason Smith * Chair of the House Republican Policy Committee: Gary Palmer * Chair of the National Republican Congressional Committee: Tom Emmer * House Republican Chief Deputy Whip: Drew Ferguson (2) A Call for American Renewal INTRODUCTION The manifesto was released one day after the ousting of Representative Liz Cheney as chair of the House Republican Conference, and was largely seen as a reaction against the influence of Trumpism within the Republican Party. (3) List of nicknames used by Donald Trump Domestic political figures - Table-0-11 [HEADER] [COL] Nickname [COL] Personal name [COL] Notes [ROW] id="The Warmonger" [COL] The Warmonger [COL] Liz Cheney [COL] U.S. representative for Wyoming's at-large congressional district; Chair of the House Republican Conference (4) Conscience vote Practice in various countries - United States Similarly, when House Republican leadership decided not to whip votes against the second impeachment of Donald Trump, Liz Cheney—the third-highest-ranking Republican—referred to the matter as a "vote of conscience".}
    \caption{
        {\bf Input Format.}
    }
    \end{justify}
\end{figure}

\clearpage
\begin{figure}[!h]
    \small
    \sf
    \begin{justify}
\nlp{\RaggedRight(2) 
Elizabeth Lynne Cheney ( ; born July 28, 1966) is an American attorney and politician who has served as the U.S. Representative for since 2017. She was the Chair of the House Republican Conference, the third-highest position in the House Republican leadership. She is the third woman elected to that position after Deborah Pryce and Cathy McMorris Rodgers. She held several positions in the U.S. State Department during the George W. Bush administration, notably as Deputy Assistant Secretary of State for Near Eastern Affairs and Coordinator for Broader Middle East and North Africa Initiatives. She promoted regime change in Iran while chairing the Iran Syria Policy and Operations Group with Elliott Abrams. In 2009 Cheney and Bill Kristol founded Keep America Safe, a nonprofit organization concerned with national security issues that advocated the positions of the former Bush administration. She was a candidate for the 2014 election to the U.S. Senate in Wyoming, challenging three-term incumbent Mike Enzi, before withdrawing from the race. In the House of Representatives, she holds the seat her father held for a decade, representing Wyoming from 1979 to 1989. Cheney is a neoconservative. She later supported the second impeachment of Donald Trump for his role in the 2021 storming of the U.S. Capitol.
}
    \caption{
        {\bf T5 Output Format.}
    }
    \end{justify}
\end{figure}

\begin{figure}[!h]
    \small
    \sf
    \begin{justify}
\nlp{\RaggedRight(2) 
[0] [1] [2] [3] [4] [5] [6] In the House of Representatives, she holds the seat that was held by her father from 1979 to 1989. (6) She is known for her neoconservative foreign policy views, and her affiliation with the Trump campaign. (0) (1) (2) (3) (4)  Cheney is under fire for her role in the second impeachment of Donald Trump in January 2021.
}
    \caption{
        {\bf \model Output Format.}
    }
    \end{justify}
\end{figure}

\clearpage
\onecolumn
\section{More Qualitative Examples}
\label{appendix:examples}
\begin{figure*}[!h]
    \centering
    \includegraphics[width=\textwidth]{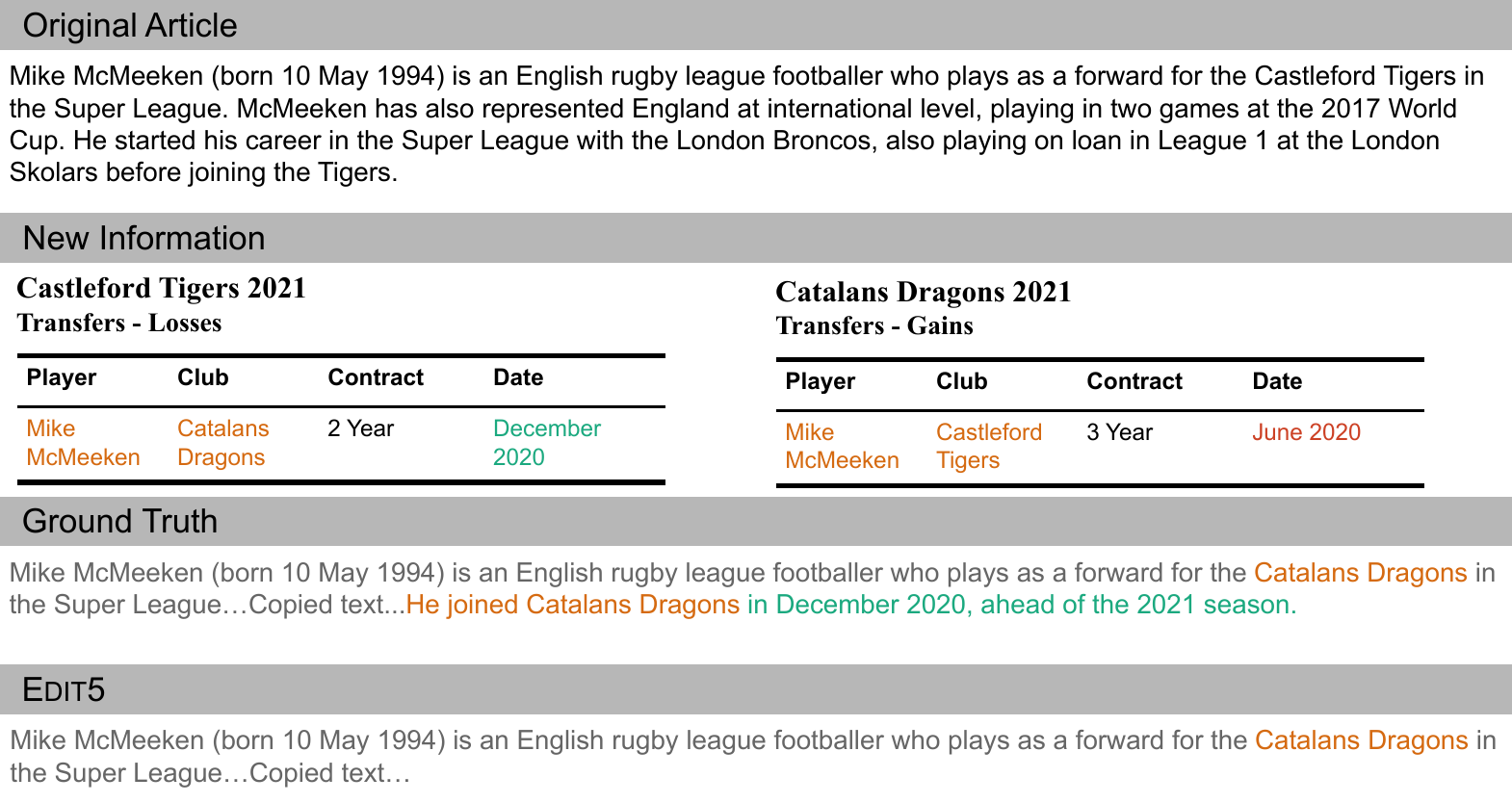}
    \caption{{\bf Example 1.}}
    \label{fig:additional-1}
\end{figure*}

\begin{figure*}[!h]
    \centering
    \includegraphics[width=\textwidth]{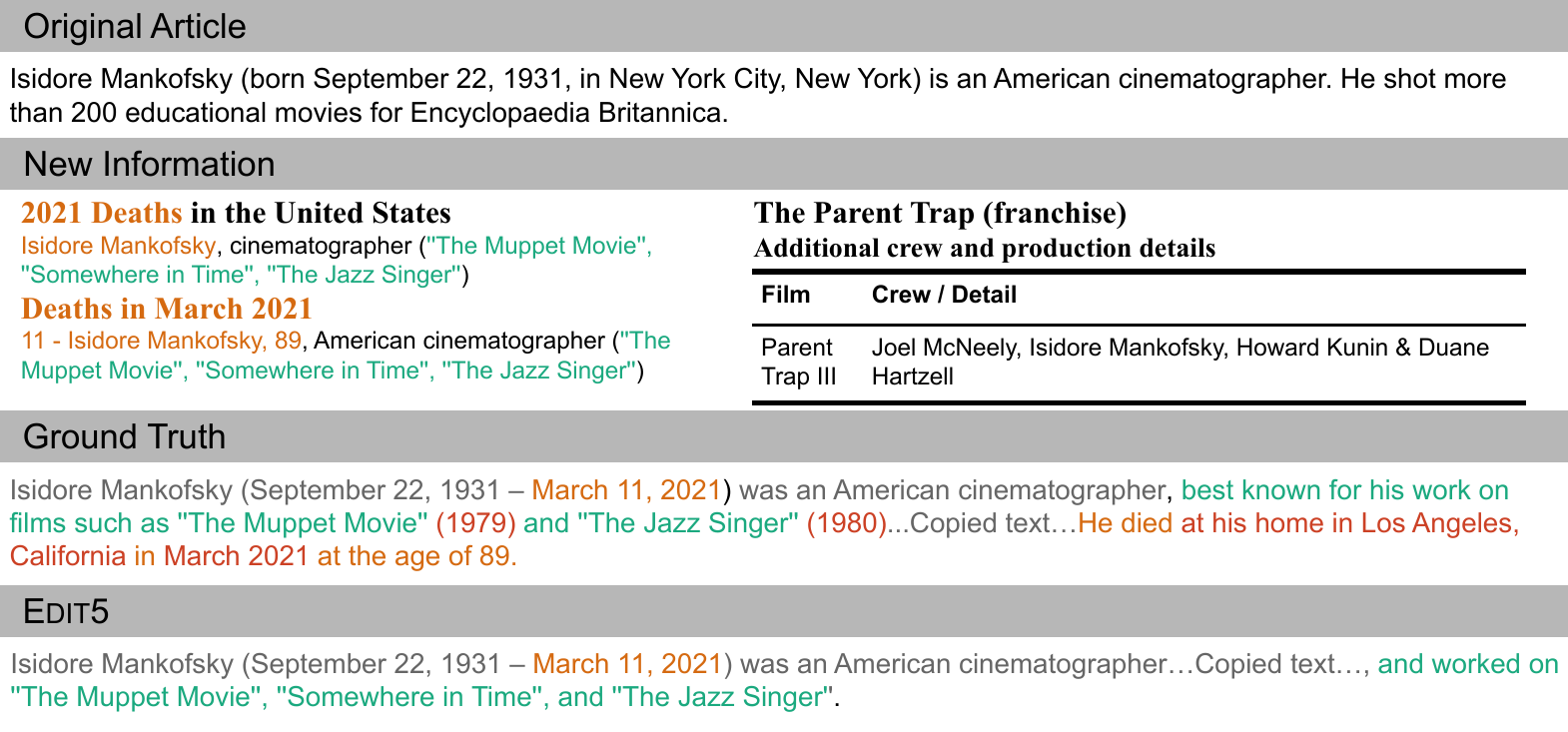}
    \caption{{\bf Example 2.}}
    \label{fig:additional-2}
\end{figure*}

\begin{figure*}[!h]
    \centering
    \includegraphics[width=\textwidth]{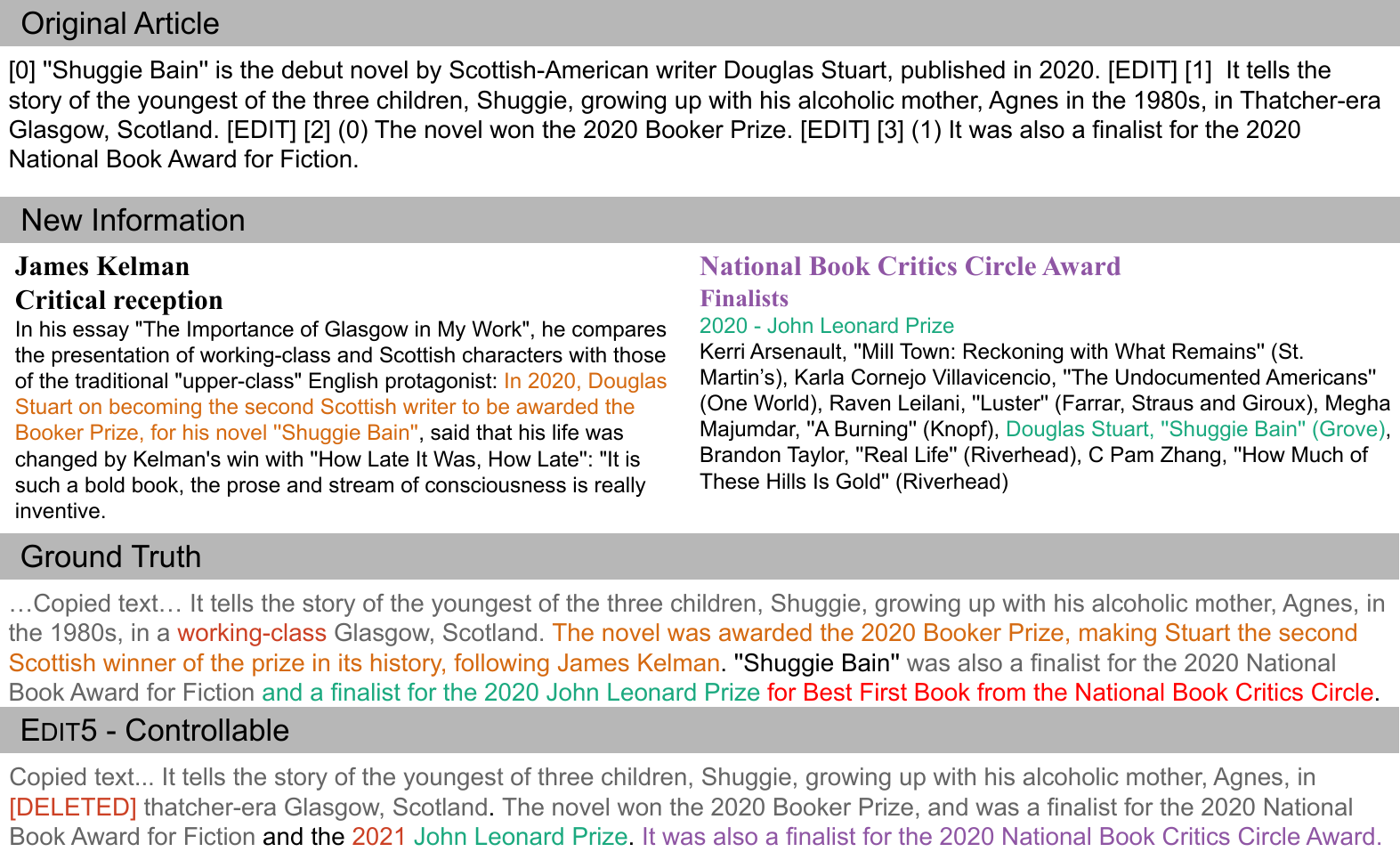}
    \caption{{\bf Using Control Codes.}}
    \label{fig:controllable}
\end{figure*}

\clearpage
\onecolumn
\begin{sidewaysfigure}[!ht]
    \centering
    \includegraphics[width=0.9\textwidth, height=0.9\textheight, keepaspectratio]{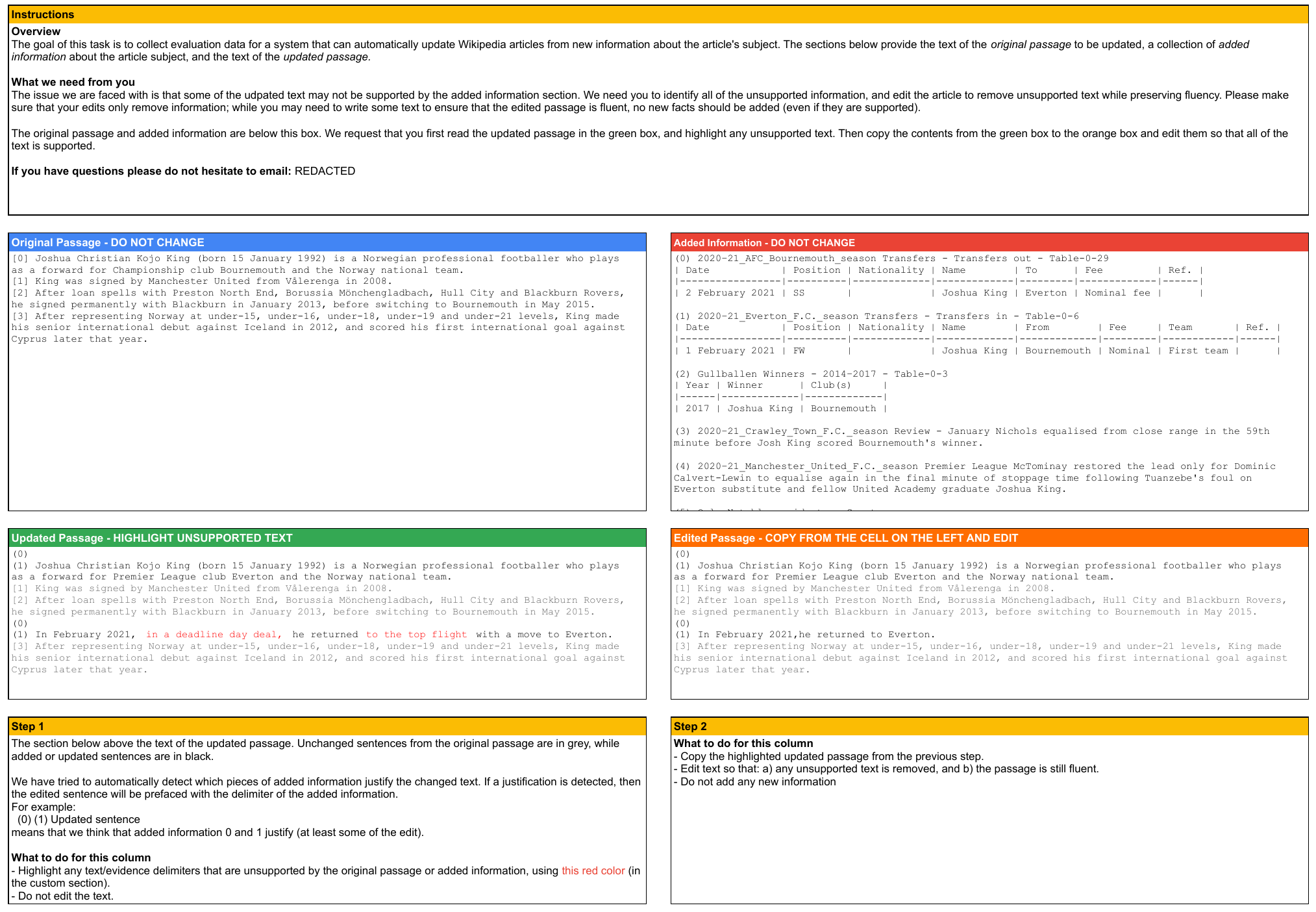}
    \caption{Annotator Interface}
    \label{fig:annotator-interface}
\end{sidewaysfigure}
\clearpage

\end{document}